# CausalKG: Causal Knowledge Graph
## Explainability using interventional and counterfactual reasoning

Utkarshani Jaimini
Artificial Intelligence Institute
University of South Carolina

Amit Sheth
Artificial Intelligence Institute
University of South Carolina

Humans use causality and hypothetical retrospection in their daily decision-making, planning, and understanding of life events [1]. The human mind, while retrospecting a given situation, think about questions such as "What was the cause of the given situation?", "What would be the effect of my action?", "What would have happened if I had taken another action instead?", or "Which action led to this effect?". The human mind has an innate understanding of causality [15]. It develops a causal model of the world, which learns with fewer data points, makes inferences, and contemplates counterfactual scenarios [8]. The unseen, unknown, scenarios are known as counterfactuals [2].

According to Gary Marcus and Judea Pearl, there is a need for Artificial Intelligence (AI) systems to have an built in understanding of causality and the ability to reason about counterfactuals [3,4]. The current KGs, such as ConceptNet and CauseNet, represent causality as a simple binary relation.

AI algorithms use a representation based on knowledge graphs (KG) to represent the concepts of time, space, and facts. A KG is a graphical data model which captures the semantic relationships between entities such as events, objects, or concepts. The existing KGs represent causal relationships extracted from texts based on linguistic patterns of noun phrases for causes and effects as in ConceptNet and WordNet. A KG represents causality as *hasCausal*, *causes*, and *mediator* relationships between cause and effect entities [9,10]. The KGs should model causality in terms of entities and not just noun phrases as in Wikidata and DBpedia. The entity-based representation model enables broader search space by linking a causal entity to relevant effect entities or concepts in KG. Causality is a complex relationship that cannot be expressed as a single link between the cause and effect entities as represented in the current KGs. The current causality representation in KGs makes it challenging to support counterfactual reasoning. A richer representation of causality in AI systems using a KG-based approach is needed for better explainability, and support for intervention and counterfactuals reasoning, leading to improved understanding of AI systems by humans.

> Representation of causality in AI systems using knowledge-graph based approach is needed for better explainability, and support for intervention and counterfactuals, leading to improved understanding of AI systems by humans.

The causality representation requires a higher representation framework to define the context, the causal information, and the causal effects, as shown in Figure 1. We show that the integration of Bayesian causality representation with KG enables counterfactual-based reasoning for better explainable and human-understandable models.

Integrating causal knowledge with the observational data can enable the machine learning models to learn the causal relationships behind the inaccurate conclusions, eventually improving the model performance. The proposed Causal Knowledge Graph (CausalKG) framework, as shown in Figure 2, leverages recent progress of causality and KG towards explainability. CausalKG intends to address the lack of a domain adaptable causal model and represent the complex causal relations using the hyper-relational graph representation in the KG. Furthermore, we show that the CausalKG's interventional and counterfactual reasoning can be used by the AI system for the domain explainability. The CausalKG utilizes the 1) domain knowledge embedded in the KG to provide a comprehensive search space for possible interventional and counterfactual variables that otherwise might be missed with just a Bayesian causality representation, and 2) the expressivity of KG to generate human understandable explanations.

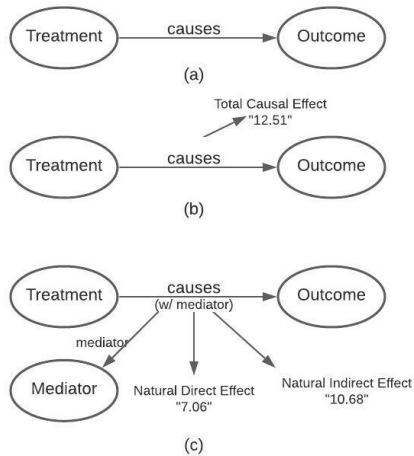

**Figure 1. (a)** Causal Representation as a single cause-effect relation. **(b,c)** Causality as a complex representation of causal effect associated with the different pathways. In (b), the total causal effect is associated between a treatment and an outcome, e.g. the treatment variable has a total causal effect of 12.51 on the outcome. In (c), the natural direct and indirect effects are associated with a mediator variable which acts as an intermediary between a treatment and an outcome. The natural direct effect is the effect of *treatment* on the *outcome* in the presence of the *mediator,* where the mediator is under the control (treatment = 0). The natural indirect effect is the effect of *treatment* on the *outcome* in the presence of the *mediator,* where the mediator receives a hypothetical change in the treatment variable (treatment =1). The treatment variable in the presence of a mediator has a natural direct effect of 7.06 and natural indirect effect 10.68 on the outcome.

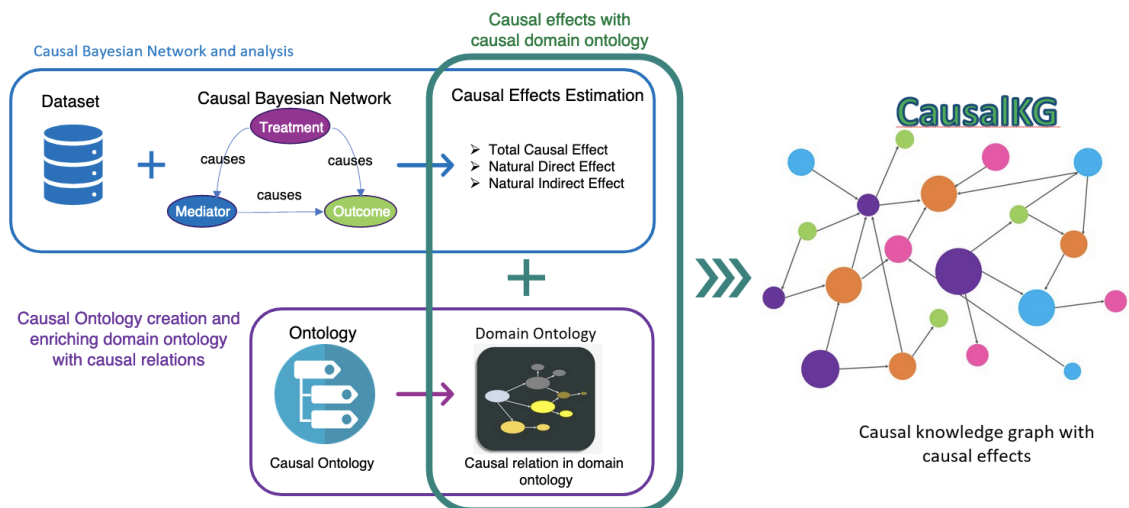

**Figure 2.** Causal Knowledge Graph Framework consists of three main steps, i) a Causal Bayesian Network and a domain-specific observational dataset, ii) Causal Ontology creation and enriching the domain ontology with causal relationships, and iii) Estimating the causal effects of the treatment, mediator, and outcome variable in the domain for a given context

> **Explainability**
>
> Explainability is the notion that a machine learning (ML) model and its results can be explained in a way that "makes sense" to a human being. The traditional machine learning algorithm tends to be more explainable but less performant. In comparison, the deep learning algorithms are more performant but are much harder to explain. We are concentrating on the explainability focused on helping users develop trust in the AI system and take action. Explainability falls under the following two categories:
>
> **Model-based explainability:**
> *Model-based Explainability* is a broad concept of analyzing and understanding the results provided by ML models. It is most often used in the context of "black-box" models. It is not easy to demonstrate how the model arrived at a specific decision. For example, a healthcare model predicts whether a patient is suffering from a disease or not. The healthcare provider needs to know what features the model is considering to determine the prediction result. There are existing tools for the black box models which assist in explaining a given result, such as feature importance, SHAP, LIME, etc.
>
> **Domain explainability:**
> *Domain explainability* is defined as providing human understandable explanations for the results generated by the ML models. For example, an explanation could be an answer to the "Why", "What if" questions that help the human understand the cause of a prediction result and gain trust in the model. Data and statistics are not enough for human understandable models. The external domain knowledge is needed to model and perform the intervention and counterfactual reasoning. For example, in the above disease prediction model, a medical guideline is provided as external domain knowledge to generate a causal explanation for the results.
>
> Explainability henceforth is referred to as domain explainability.

## Causal Knowledge Graph

A CausalKG is a hyper-relational graph. It is used to represent causality as a complex relation in the graph, which may involve more than two entities and may be annotated with additional information such as causal effect. The edge connected with more than two nodes represents the causal relationship, mediator variable, and the associated causal effects. The mediator is a variable which causes mediation between the Treatment (an independent variable) and Outcome (a dependent variable). It also explains how different intervention effects the outcome variable. The associated causal effects can be of three types: total causal effect, natural direct effect, and natural indirect effect. A CausalKG explicitly takes the causal knowledge into account, e.g., allowing flexible incorporation of the causal domain knowledge represented by a causal bayesian network, and automating the causal inference tasks. Thus, this approach blends causal Bayesian network, causal ontology design, and knowledge representation described below.

The goal of CausalKG is to support the integration of causal knowledge into the KGs for improving domain explainability, promoting interventional, counterfactual reasoning and causal inference in downstream AI tasks. These capabilities cannot be achieved if causality is based only on the Bayesian causality representation, and observational data, but the causal knowledge encapsulated in the KG makes that possible.

## Causal Bayesian Network

Judea Pearl introduced the causal Bayesian network (CBN) as a representation tool for causality. Similar to a Bayesian network, a causal Bayesian network is a directed acyclic graph G = <V, E>, where nodes V represent causal variables and edges E represent the conditional causal influences among the causal variables [3]. A CBN is a graphical representation that expresses causal knowledge in a given domain. The graphical representation is intuitive and human-readable. A CBN assists in understanding the causal relations, supporting the ability to represent and respond to changes in the causal system. The graphical representation of causal knowledge allows access to the quantitative prediction of the effect of an intervention on one or more variables using total causal effect, natural direct effect, and natural indirect effect. The joint distribution in the Bayesian network is the probability of events and how the probabilities will change given a set of observations. A CBN evaluates how the probability would change due to external intervention and estimates the effect of an intervention. This approach is extensively used for policy analysis, planning, and medical treatment management [11]. Each possible intervention has a new causal model, C. A causal model cannot be interpreted in a standard propositional logic or probability calculus because it deals with changes in the counterfactual world (out-of-distribution) rather than changes in our beliefs about observational data.

**Causal Ontology**
The Causal Ontology is a taxonomical representation of domain facts and relationships between entities that represents a generic representation of a causal model. The three causal classes are *Treatment*, *Mediator*, and *Outcome* with causal relationships *causes* and *causesWith*. In addition, total causal effect, natural direct effect, and natural indirect effect are represented as data properties in this ontology. The designed ontology can be extended to a given domain ontology to describe the causal relationships between the domain entities. As an initial input, the framework requires a CBN describing the causal relationships of the domain and the domain ontology with causal relations extended using the causal ontology.

**Causal Knowledge Representation**
A CausalKG utilizes the benefits of CBNs, causal ontology, and KGs to provide robust and explainable insights. The KGs have often used the resource description framework (RDF) to represent information on the web. RDF consists of triples or statements with subject, predicate, object. RDF captures binary entities, connecting two objects, or the relationship between two entities. RDF uses reification to represent n-ary relations, i.e. relations linking more than two entities. Reification creates an intermediate node which groups two or more entities, making the RDF complex and less intuitive. RDF* (pronounced as RDF-star) is an extension to RDF that treats a triple as a single entity using nested or embedded triples. An entire triple can become the subject of a second triple, which enables assigning metadata and attributes to a triple. CausalKG uses RDF* to represent complex causal relationships (as shown in Figure 6).

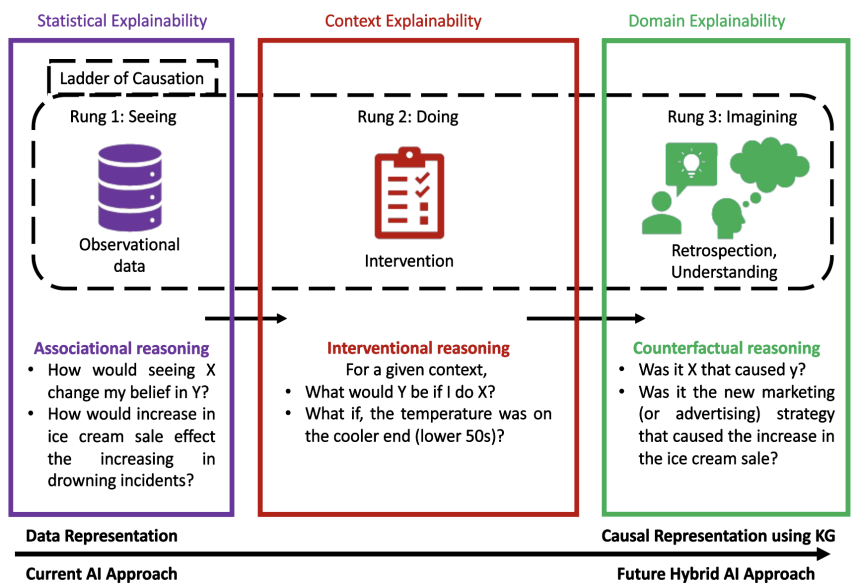

Figure 3. Causal Artificial Intelligence: From statistical explainability based on data representation and associational support to context explainability based on causal relation and interventional support eventually leading to domain explainability based on causal representation in KG leading to counterfactual support

**Explainability**
There is a gap between the observation-based and counterfactual-based explainability in AI. Figure 3 illustrates the future of AI systems transforming from purely statistical (i.e., associational) explainability based on the observational data to context explainability (i.e., interventional), eventually leading to domain explainability involving counterfactuals and causal reasoning. We emphasize the need for explainability in AI that goes beyond explaining the algorithmic inner working to also provide human-understandable explanations necessary for users to trust a given AI model using three categories of explainability - statistical, context, and domain explainability.

**Statistical Explainability**
Statistical explainability generates an explanation for the co-occurrence of a given phenomenon based on the statistical (or associational) methods such as correlations [6]. The statistical explainability may fail to capture the underlying relationship between the observational data, which can lead to inaccurate conclusions (such as spurious correlations). For example, the statistical pattern of higher ice cream sales and a higher number of people drowning leads to the incorrect conclusion that increasing ice cream sales leads to an increase in people drowning in water. The above result is analyzed out of context, which disregards the hidden causal knowledge leading to an inaccurate conclusion.

**Context Explainability**
Context explainability is a means to generate a human-understandable explanation taking the context information of a given observation into account [5]. The inaccurate conclusions are a result of out-of-context analyses of the relationship between entities. The context information is at times not present in the observational data. However, it can be deduced from the underlying causal variables called confounders (also confounding variables). A given context can be represented using a causal ontology and a CBN. The causal ontology gives us the causal relationship between the entities for a given context. The CBN presents a graphical representation of a given context enabling *interventional reasoning* (e.g., "What if I had taken action *B* instead?"). In the above example, the statistical results need to be analyzed in the context of the weather or the temperature. The higher ice cream sales and the higher number of people drowning are both due to the rise in temperature during the summer months. As the temperature rises during the summer months, people are out in the water leading to water-related accidents. The hike in temperature also shows the increase in ice cream sales to relieve the high temperature. The causal knowledge of temperature affecting the ice cream sales, and water-related accidents can be captured using a KG and can assist in pointing out the underlying cause to both the consequences. A CBN constructed using the above context information enables interventional reasoning such as "What if the temperature was cold? How would it affect the ice cream sale and the number of people drowning?", "What if the temperature was hot? How would it affect the ice cream sale and the number of people drowning?"

**Domain Explainability**
Domain explainability explains the underlying causal relations using observational data, domain knowledge, and counterfactual reasoning. Domain explainability considers existing domain knowledge and causal relations between entities. David Hume, a philosopher, and empiricist analyzed causality in terms of sufficient and necessary conditions, A causes B such that A is necessary and sufficient cause for B [7]. As defined by Hume, causality can be achieved using domain explainability accounting for the sufficient and necessary conditions for causality. The CausalKG, with the causal relations, enables domain explainability using counterfactual reasoning ("e.g., Was it action A which led to this effect?").

The above example can be explained in terms of a given domain using counterfactual reasoning such as "Was it the new marketing strategy that caused the increase in the ice cream sale?", "Was it the lack of supervision near the pool that caused the increase in drowning?". CausalKG provides the context to consider necessary and sufficient conditions needed to improve domain explainability using counterfactual reasoning.

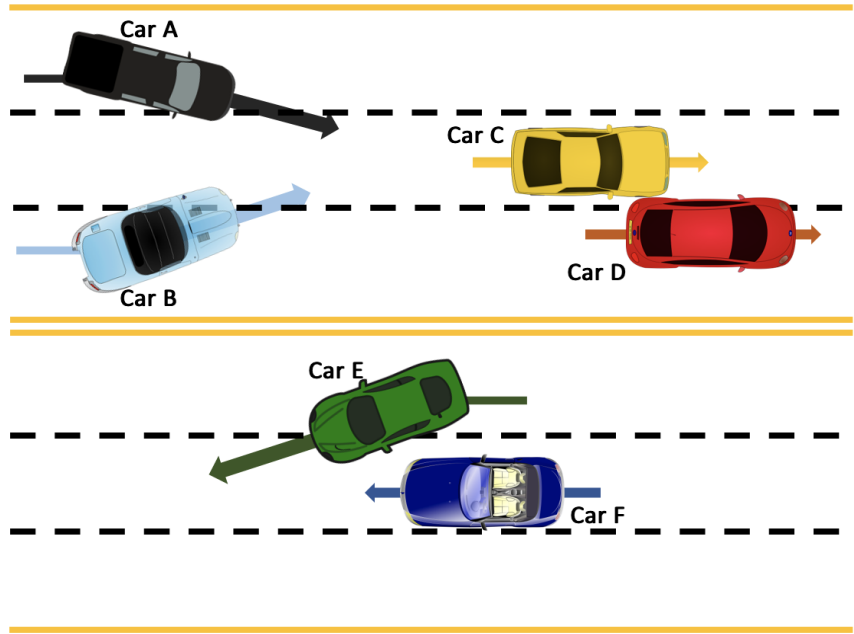

**Figure 4.** Highway Collision scene, a snapshot of possible scenarios on a highway that might lead to a collision between two vehicles when either i) the driver fails to identify the passing vehicle in the nearby lane due to distraction or blind spots, or ii) loses control of the vehicle due to slippery road.

### Use case: Autonomous Driving

AI systems and domain problems require an accurate representation of cause(s) and their effect(s). Therefore, it is not sufficient solely to understand the cause(s) of an outcome but the underline effect(s) and mediator(s), if any. We present a use case in autonomous driving (AD) to explain how the CausalKG for causal representation supports domain explainability.

Tesla customers can now request a fully self-driving car equipped with the beta version of Tesla's advanced driver-assistance suite (https://wapo.st/3pfx8FB). Driving is a complex system that requires meticulous planning and vigilance during execution. A human brain, while driving, analyses the environment based on the observation and, at the same time, retrospects and prospects possible scenarios and the corresponding measures to be taken. While AD systems are well trained on observational data, they perform poorly in unseen, uncertain, and risky driving scenarios [12].

The AD vehicle works in parallel with other entities in the driving scenes, such as pedestrians, traffic signs, and other vehicles. A representation of the relationship and interaction of the entities within a driving scene is essential for better understanding the AD vehicle's behavior in a given context. Some of the interventional and counterfactual reasoning which can assist in AD understanding is- "How does the entities in the driving scene effects the autonomous driving vehicle?", "What if a pedestrian is jay-walking; how would it effect the vehicle's behavior (i.e., stop or keep moving)?", "What if the vehicle fails to identify the stop line marking? How would it effect the vehicle's behavior concerning a pedestrian?" and "How does the sudden lane change by a vehicle effect the adjacent vehicle?".

According to the National Highway Traffic Safety Administration (NHTSA), in 2019, there were 3142 fatal accidents due to distracted driving, 697 accidents due to drowsy driving, 718,000 accidents due to rain (2,569 of which were fatal), and 182,000 accidents due to snow (440 of which were fatal) (https://bit.ly/3D8cWdS). The US Department of Labor, Mine Safety and Health Administration Safety Manual No. 10 is used for accident investigation (https://bit.ly/3d5kW4z). The safety manual presents a detailed analysis of an accident investigation on what, how, and why the accident happened. It reveals three cause levels of an accident: basic, direct, indirect cause, which correspond to total, natural direct, and natural indirect cause, respectively, in the CausalKG. With the fully automated AD vehicles on the road, it is crucial to understand the sequence of the events which led to an accident to devise a mitigation plan for similar situations in the future, and assist with insurance and law related proceedings. CausalKG is instrumental in the context of AD vehicles to represent and understand the causal relationships between entities in a driving scene, their behaviors, and possible hypothetical scenarios. It can further assist in interventional and counterfactual reasoning to equip AD vehicles for unseen and risky scenarios.

The typical accident scenario is a collision due to a distracted driver or slippery road, leading to sudden lane change, as shown in Figure 4. In addition, a driver could be distracted due to phone usage or be under the influence of alcohol, which might lead to sudden lane change and collision into the vehicle in the next lane. The collision due to sudden lane change can also be due to bad weather leading to slippery roads and loss of control.

Observational data can assist with
- Statistical Explainability - What is the likelihood of a vehicle crashing if the driver is distracted?
- Context Explainability - What if the driver is distracted?
- Domain Explainability - Was the collision due to phone usage or the slippery roads?

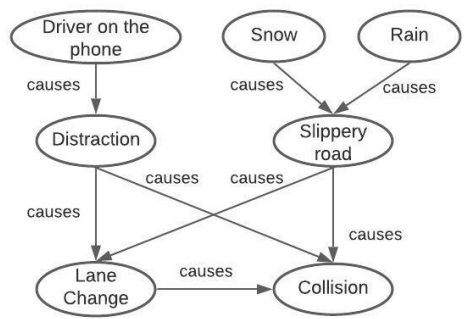

**Figure 5. A Collision Bayesian Network, a graphical representation of the causal relations and interaction between entities in collision scenarios.**

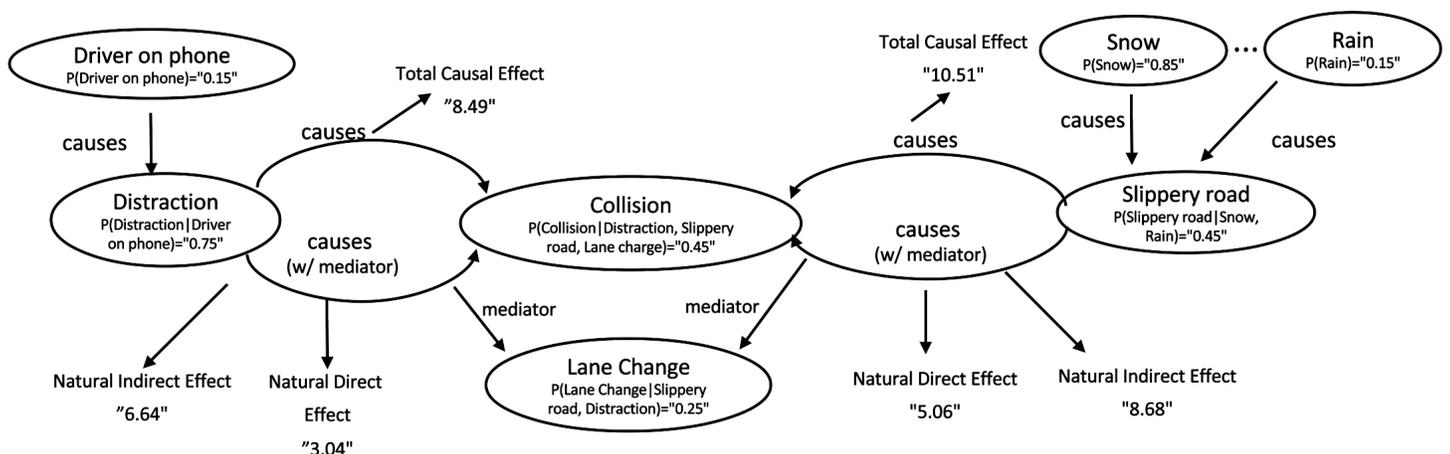

**Figure 6. A snapshot of CausalKG for collisions due to driver distraction or slippery road scenarios. Each node in the CausalKG is a concept in KG and is associated with a conditional probability estimated using the CBN in the previous step (as shown in Figure 2). The edges between the nodes represent the causal relationships between the concepts.**

The above collision scenario can be graphically represented as a CBN expressing causal relations between the driving scene entities. In Figure 5, the distraction of a driver (e.g., caused due to cell phone usage) or slippery roads (e.g., caused due to snow, or rain) causes a sudden lane change, leading to a collision. These variables are dependent on other observed variables in the system and are known as endogenous (dependent) variables. The external variables such as cellphone usage, driving under the influence of alcohol, weather conditions such as snow and rain are not affected by other variables in the system and are known as exogenous (independent) variables. The existing driving scene ontology is enriched with the causal relation from the causal ontology to capture the relationships between the entities, such as distracted driving (or slippery road) causes a sudden lane change and collision; and a sudden lane change causes a collision into the nearby vehicle [14].

The different types of causal effects (i.e., total causal effect, natural direct effect, and natural indirect effect) are estimated using the autonomous driving dataset for the above CBN. They express the quantitative effect of the various interventions on the AD entities and aid in explainability using interventional and counterfactual reasoning. The generated CausalKG shown in Figure 6 captures the causal relations between the AD entities and their quantitative effects. The CausalKG demonstrates the explainability in the AD using the counterfactual and interventional reasoning listed below. Such representation supports a better understanding of the behavior of entities in the driving scene in an unseen, risky situation.

- **Total Causal Effect (Basic cause):** How would the driver's distraction (or slippery road) effect the occurrence of a vehicle collision?
- **Natural Direct Effect (Unplanned or direct cause):** What if the vehicle fails to identify the passing vehicle in the adjacent lane (what if there is a blind spot), how would it effect the vehicle's collision? In this scenario, the lane change is not due to the distraction (or slippery road), but the collision is due to the distraction.
- **Natural Indirect Effect (Unsafe act or indirect cause):** What if, there is a lane change due to the distraction (or slippery road), how would it effect the vehicle's collision? In this scenario, the lane change is due to distraction (or losing control over the vehicle under risky situations). However, the collision is not due to distraction.

**Knowledge-infused learning using a CausalKG**

CausalKG is a step towards symbolic AI for causality-based knowledge-infused learning [13]. CausalKG AI systems do not learn solely from correlations but, instead, have a causal representation of the world around them. Apart from causality, CausalKG, with the help of a KG, also represents time, space, physical objects, and humans and their interactions.

Current AI algorithms rely on independent and identically distributed data. As a result, they cannot infer out-of-distribution or hypothetical, interventional scenarios. CausalKG can be used to infuse existing KGs with causal knowledge of the domain to enable interventional and counterfactual reasoning that can be deduced using observational data and domain expert knowledge. The advantage of constructing a CausalKG is the integration of causality in reasoning and prediction processes, such as the agent action understanding, planning, medical diagnosis process, etc. Such integration can improve the accuracy and reliability of existing AI algorithms by providing better explainability of the outcome.

### Acknowledgements

This work has greatly benefited from the discussions with Dr. Cory Henson. We also thank Dr. Christian O'reilly and Ruwan Wickramarachchi for their in-depth review. This research is support in part by National Science Foundation (NSF) Award # 2133842 "EAGER: Advancing Neuro-symbolic AI with Deep Knowledge-infused Learning," and Award #2119654, "RII Track 2 FEC: Enabling Factory to Factory (F2F) Networking for Future Manufacturing."
Any opinions, findings, and conclusions or recommendations expressed in this material are those of the author(s) and do not necessarily reflect the views of the NSF.

**Authors**

Utkarshani Jaimini is a Ph.D. student at the Artificial Intelligence Institute, University of South Carolina. Her dissertation research is focused on developing a richer representation of causality using knowledge graphs for better explainability with the applications in autonomous driving and healthcare. Contact her at ujaimini@email.sc.edu

Amit Sheth is the Founding Director of Artificial Intelligence Institute, University of South Carolina. His current core AI research is in knowledge-infused learning and explanation, and translational research includes personalized and public health, social good, education, and future manufacturing. He is a fellow of IEEE, AAAI, AAAS, and ACM. http://aiisc.ai/amit